\documentclass[conference]{IEEEtran}
\IEEEoverridecommandlockouts

\usepackage{pgfplots}

\usepackage{cite}
\usepackage{amsmath,amssymb,amsfonts}
\usepackage{algorithmic}
\usepackage{graphicx}
\usepackage{algorithm,algorithmic}
\usepackage{hyperref}
\hypersetup{hidelinks=true}
\usepackage{textcomp}

\usepackage{comment}
\usepackage{booktabs}
\usepackage[table]{xcolor} 
\usepackage{colortbl} 

\definecolor{low}{RGB}{198, 239, 206}    
\definecolor{medium}{RGB}{255, 235, 156} 
\definecolor{high}{RGB}{255, 199, 206}   

\usepackage{cite}
\usepackage{amsmath,amssymb,amsfonts}
\usepackage{algorithmic}
\usepackage{graphicx}
\usepackage{textcomp}
\usepackage{xcolor}
\usepackage{subfigure}
\usepackage{caption, subcaption}
\usepackage{amsmath,graphicx}
\usepackage{textcomp}
\usepackage{amsmath,amssymb,amsfonts}
\usepackage{mathtools}
\usepackage{amsmath,graphicx, multirow, hyperref}
\usepackage{subcaption, subfigure}
\usepackage{graphicx,subfigure}

\newcommand{\cognospeak}{CognoMemory}
\newcommand{\cognospeakOLD}{CognoSpeak}
\newcommand{\dementiabank}{DementiaBank}
\newcommand{\bart}{BART}

\newcommand{\roberta}{RoBERTa}
\newcommand{\distilbert}{DistilBERT}

\newcommand{\egemaps}{eGeMAPS}
\newcommand{\compare}{ComParE}

\newcommand{\bow}{BoW}
\newcommand{\tfidf}{TF-IDF}

\newcommand{\WV}{Wav2Vec 2.0}
\newcommand{\Nemo}{NeMo}

\usepackage{pgfplotstable}
\usepackage[table]{xcolor}
\usepackage{booktabs}
\usepackage{colortbl}
\usepackage{multirow}
\usepackage{caption}
\usepackage{tikz}
\pgfplotsset{compat=1.18}

\newcommand{\minWER}{0.3}
\newcommand{\maxWER}{0.65}

\newcommand{\heatmapcell}[1]{%
    \pgfmathsetmacro{\value}{#1}
    \pgfmathsetmacro{\percent}{100*(\value - \minWER)/(\maxWER - \minWER)}
    \edef\cellcolorval{\noexpand\cellcolor{red!\percent!white}}%
    \cellcolorval #1%
}

\newcommand{\heatmapcellPVALs}[1]{%
    \pgfmathsetmacro{\value}{#1}
    \pgfmathsetmacro{\percent}{100*\value}
    \ifdim \percent pt < 5 pt
        \pgfmathsetmacro{\setCellColor}{"green!30!white"}
    \else
        \pgfmathsetmacro{\setCellColor}{"red!50!white"}
    \fi
    \edef\cellcolorval{\noexpand\cellcolor{\setCellColor}}%
    \cellcolorval #1%
}

\newcommand{\sig}[1]{%
  #1%
  \pgfmathsetmacro{\pvalue}{#1}%
  \ifdim \pvalue pt < 0.001 pt\relax$^{***}$%
  \else\ifdim \pvalue pt < 0.01 pt\relax$^{**}$%
  \else\ifdim \pvalue pt < 0.05 pt\relax$^{*}$%
  \fi\fi\fi%
}

\def\BibTeX{{\rm B\kern-.05em{\sc i\kern-.025em b}\kern-.08em
    T\kern-.1667em\lower.7ex\hbox{E}\kern-.125emX}}
\begin{document}

\title{False positive bias in AI-powered speech-based cognitive screening for multilingual English speakers in the UK\\
}

\author{\IEEEauthorblockN{Madhurananda Pahar$^1$\IEEEauthorrefmark{1},
Caitlin Illingworth$^3$\IEEEauthorrefmark{1},
Dorota Braun$^3$\IEEEauthorrefmark{1},
Bahman Mirheidari$^1$\IEEEauthorrefmark{1},\\
Lise Sproson$^2$\IEEEauthorrefmark{4},
Daniel Blackburn$^3$\IEEEauthorrefmark{1},
Heidi Christensen$^1$\IEEEauthorrefmark{1}
}
\vspace{5pt}
\IEEEauthorblockA{$^1$School of Computer Science, University of Sheffield, Sheffield, S1 4DP, UK\\
$^2$NIHR Devices for Dignity HTC, Sheffield Teaching Hospitals NHS Foundation Trust, Sheffield, S10 2JF, UK\\
\vspace{5pt}
$^3$Sheffield Institute for Translational Neuroscience (SITraN), University of Sheffield, Sheffield, S10 2HQ, UK\\
Email: \IEEEauthorrefmark{1}\{m.pahar, c.illingworth, d.a.braun, b.mirheidari, d.blackburn, heidi.christensen\}@sheffield.ac.uk, \IEEEauthorrefmark{4}lise.sproson@nihr.ac.uk
}
}


\maketitle

\begin{abstract}

\textit{Objective:}
Conversational speech can reveal early signs of cognitive decline, including dementia and mild cognitive impairment (MCI). While artificial intelligence (AI) models have demonstrated promising performance for speech-based cognitive screening, most studies focus on monolingual populations. In the United Kingdom (UK), dementia prevalence is projected to increase most rapidly among Black and Asian communities, where multilingualism is common. Hence, it is important to assess whether these systems operate equitably across multilingual populations.
\textit{Approach:}
We recruited 1,395 participants, including monolingual English speakers from across the UK and multilingual speakers from community centres in Sheffield and Bradford, contributing over 263 hours of speech collected using the \cognospeak{} virtual agent. Multilingual participants spoke English alongside Somali, Chinese, or South Asian languages (e.g., Hindi, Urdu, Punjabi, Mirpuri, Arabic). We evaluated automatic speech recognition (ASR) performance and examined downstream disparities in AI models performing cognitive status classification and Mini-Mental State Examination (MMSE) score regression.
\textit{Results:}
ASR systems (Whisper, \WV{}, and \Nemo{}) showed no statistically significant differences in transcription accuracy across language groups. However, downstream models demonstrated systematic disparities like multilingual participants were more frequently misclassified as cognitively impaired, particularly in memory, fluency, and reading tasks. False positive rates were substantially higher for multilingual speakers (28-37\%) compared with monolingual speakers (12-16\%), indicating that multilingual individuals were approximately 2.5 times more likely to receive an incorrect impairment classification. These disparities were further amplified when models were trained on the DementiaBank dataset.
\textit{Significance:}
To our knowledge, this study represents the first large-scale analysis of false positive bias in speech-based AI cognitive screening systems within multilingual ethnic minority populations in the UK. Our findings demonstrate that, despite strong overall performance, current speech-based AI models exhibit measurable disparities that disproportionately affect multilingual speakers. Addressing these biases will be essential for ensuring that AI-assisted cognitive screening tools can be safely and equitably deployed in linguistically diverse healthcare settings.

\end{abstract}


\section{Introduction}
\label{sec:intro}

Cognitive and memory difficulties may arise from factors such as fatigue, stress, and illness, often intensifying with age. When persistent, these symptoms may indicate mild cognitive impairment (MCI) \cite{davis2018estimating}, a condition involving deficits in memory, learning, attention, reasoning, language, and motivation \cite{rosenberg2013association}. Approximately half of individuals with MCI progress to dementia, the most common of which is Alzheimer’s disease (AD) 
\cite{prestia2013prediction, knopman2003essentials, thabtah2020correlation, hendrie1998epidemiology}. 
While early detection facilitates interventions that may delay disease progression \cite{robin2021using}, current diagnostic approaches, including neurological evaluations, historical cognitive assessments \cite{shi2023speech}, magnetic resonance imaging (MRI), standardised testing, and cerebrospinal fluid analysis \cite{yang2022deep}, are resource-intensive and unsuitable for large-scale screening \cite{mckhann2011diagnosis}. Consequently, there is a growing need for remote, intelligent technologies to enhance diagnostic efficiency and support healthcare delivery \cite{gauthier2021world}.

Speech and language impairments are well-established hallmarks of AD and can help differentiate between subjective and pathological cognitive decline \cite{kaltsa2024language}. 
Features such as reduced lexical diversity, increased repetition, simplified language, and prolonged pauses have been linked to early stages of decline and underlying tau pathology \cite{mueller2018connected, young2024speech}. 
Importantly, these linguistic markers provide a non-invasive and easily obtainable indicator of cognitive status, which can be automatically analysed using state-of-the-art artificial intelligence (AI) methods without the need for direct clinician involvement \cite{pan2021using, mirheidari2023identifying}.

For automatic systems to be clinically effective in healthcare environments such as the National Health Service (NHS) in the United Kingdom (UK), they must perform reliably across diverse demographic groups, including ethnic minority populations. Dementia prevalence is projected to increase most rapidly among these communities \cite{zhang2016life, yaffe2013effect, steenland2016meta, monica2021alzheimer}, where multilingualism is common and English is often spoken as an additional language \cite{cencus2021}. Individuals in these populations frequently face additional barriers to dementia diagnosis \cite{tsai2024assessment}, contributing to the underrepresentation of Black and South Asian patients in NHS memory assessment services \cite{teager2024retrospective}. These disparities highlight an important limitation of existing AI-based cognitive screening systems, which have rarely been evaluated across multilingual populations.

We have recently developed \cognospeak{} \cite{cognomemory_website}, formerly \cognospeakOLD{} \cite{blackburn2023developing}, an innovative online, language-based screening tool for the at-home detection of cognitive decline. By leveraging both acoustic and linguistic features, as demonstrated in our previous work \cite{pahar2025cognospeak, pahar2025mutlimodalfusion, yaointerspeech26, young2025can, pahar2025cognospeakWiley, illingworth2025developing}, \cognospeak{} is designed to accelerate the identification of early cognitive impairment. The clinical utility of \cognospeak{} critically depends on its diagnostic accuracy across diverse populations, particularly multilingual individuals within ethnic minority groups.
In this pilot study, we therefore examine the fairness and reliability of AI models within \cognospeak{}, with a particular focus on identifying and quantifying false positive bias between monolingual and multilingual participants in the UK.
To address this overarching research objective, we investigate the following research questions:
1) Do automatic speech recognition (ASR) systems perform comparably for monolingual and multilingual speakers?
2) Do classification models produce comparable false positive rates (FPR) across monolingual and multilingual speakers?
3) Are predicted cognitive scores, such as the Mini-Mental State Examination (MMSE), systematically different for multilingual speakers?
4) What insights can qualitative linguistic analysis provide regarding the presence and nature of such bias?

\begin{figure*}[h]
  \centering 
  \includegraphics[width=\linewidth]{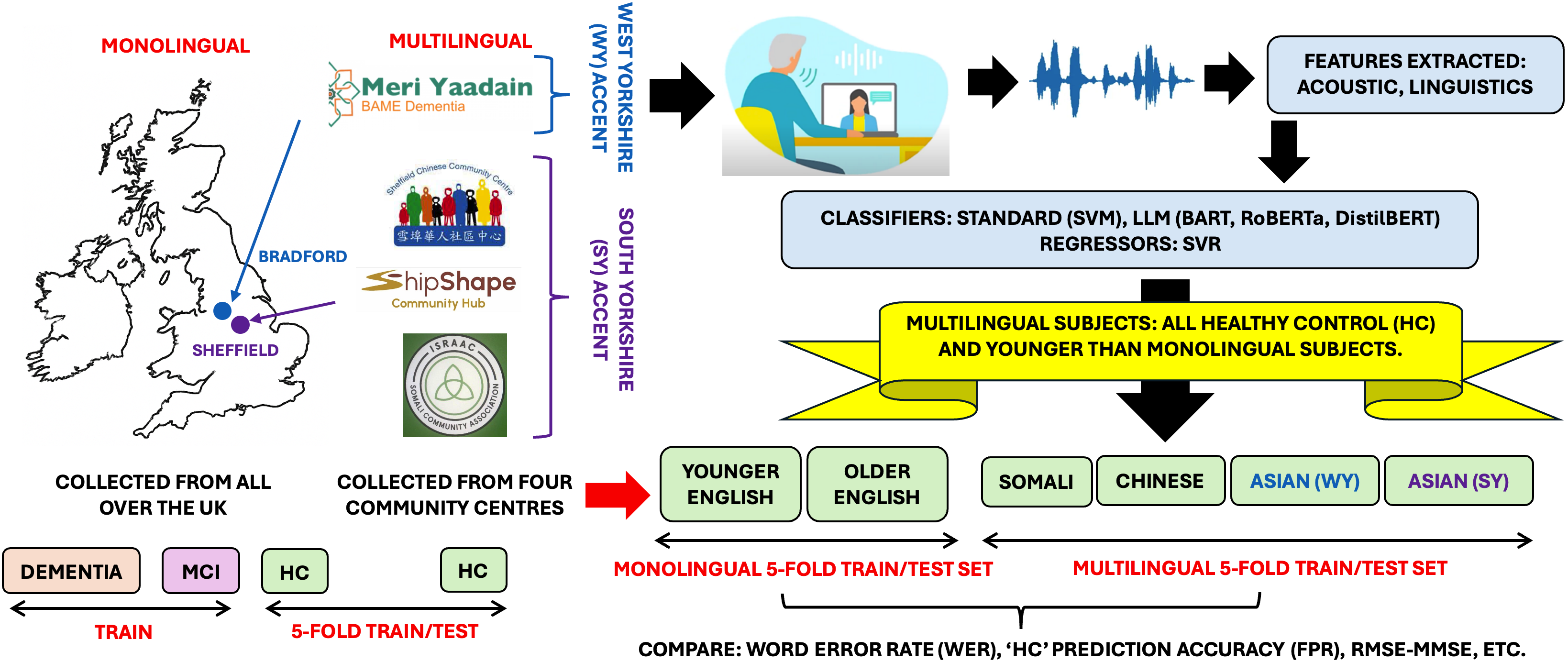} 
  \caption{\textbf{Data collection and testing summary:} Monolingual speakers came from all over the UK, whereas multilingual speakers were recruited from the four community centres in Sheffield and Bradford: the Sheffield Chinese Community Centre (Mandarin and Cantonese Chinese-English multilinguals), ISRAAC (Somali-English multilinguals), ShipShape  (Sheffield South Asian- English multilinguals), and Meri Yaadain (Bradford South Asian-English multilinguals). South Asian languages included Hindi, Urdu, Punjabi, Mirpuri and Arabic. All participants undergo \cognospeak{} assessments, answering 14 memory-probing, clinically effective question prompts asked by a virtual agent. These question prompts are used to extract both acoustic and linguistic features, which are then used to train classifiers and regressors for comparison between monolingual and multilingual speakers, as described in Table \ref{table:dataset}. } 
  \label{fig:summary}
\end{figure*}

The contributions of this study lie in addressing the research questions outlined above and in our efforts to recruit participants from ethnically diverse and multilingual communities in the UK, thereby enhancing the representativeness of the \cognospeak{} dataset. 
To this end, we collaborated with four ethnic minority community centres in two cities from the UK (Figure \ref{fig:summary}), recruiting participants from Somali, Chinese, and South Asian groups in Sheffield, where speech was characterised by a South Yorkshire accent, as well as from a South Asian community in Bradford, representing the distinct West Yorkshire dialect commonly observed in both British and British Asian speakers \cite{watt2001spectrographic, kirkham2017ethnicity}. 
Moreover, studies confirm the existence of unique linguistic features among the multilingual South Asian speakers with West Yorkshire accents \cite{rathcke2011exploring, hall2017rhotic, kirkham2015acoustic, wormald2014bradford}. Given this unique observed accent, this study also aims to facilitate better accent monitoring to distinguish between South Asian multilinguals from South Yorkshire (Sheffield) and those from West Yorkshire (Bradford).
Due to recruitment challenges and language proficiency requirements for the English-based assessment tasks, only healthy multilingual participants were available for inclusion in this pilot study.
Section \ref{sec:previouswork} reviews existing literature, followed by a summary of this novel data collection effort in Section \ref{sec:data}. 
Classifier training and evaluation for the three diagnostic categories, dementia, MCI, and healthy controls (HC), using monolingual and multilingual participants, are described in Section \ref{sec:ex_setup}  and Figure \ref{fig:summary}. 
These analyses first address the performance of ASR systems in Section \ref{subsec:wer}, where we observe visually noticeable but statistically non-significant differences in word error rates (WER) between monolingual and multilingual participants. 
Classifier performance is then compared across groups using both \cognospeak{}, the largest dataset of its kind, and DementiaBank, the second-largest publicly available dataset, which contains only binary diagnostic labels of dementia and HC. 
Here, statistically significant biases are identified in both datasets, with stronger effects observed in DementiaBank, addressing the second research question in Section \ref{subsec:classification}. 
Predicted MMSE scores further reveal statistically significant differences between monolingual and multilingual groups in Section \ref{subsec:regress}, thereby answering the third research question. 
Finally, qualitative analysis using Term Frequency-Inverse Document Frequency (\tfidf{}) metrics demonstrates that participants are more likely to refer to their places of origin, and the healthy participants also mention prominent political figures like British prime ministers while answering the memory prompts, thus addressing the fourth research question in Section \ref{subsec:qualanalysis}. 
These findings are critically discussed in Section \ref{sec:discussion}, with concluding remarks provided in Section \ref{sec:conclusion}.

\section{Previous work}
\label{sec:previouswork}

Recent advances in AI-based systems using speech technologies relying on Automatic Speech Recognition (ASR) for detecting cognitive decline in multilingual populations have emphasised system design, multilingual speech processing, and model performance. Early research predominantly focused on monolingual English datasets, but more recent studies utilise multilingual corpora such as ADReSS-M~\cite{luz2023multilingual}, TAUKADIAL~\cite{barrera2024interspeech}, and MultiConAD~\cite{shakeri2025multiconad}, which include languages like English, Chinese, Greek and Spanish. 
The performance of the ASR systems varies among native and non-native speakers \cite{hollands2022evaluating}, and while progress has been made toward developing language-agnostic ASR-based systems capable of assessing cognitive decline across diverse populations~\cite{fraser2019multilingual,cheng2024cognivoice,ye2023lamm}, these studies generally trained and tested models on different languages collected across various countries and continents. Consequently, they have not addressed cognitive decline detection in multilingual individuals from ethnic minority groups who speak the same language as native speakers due to a lack of data \cite{heng2023understanding}, a gap this study aims to fill.


Building on this motivation, we leverage the \cognospeak{} dataset, which is nearly six times larger than DementiaBank and includes 14  question prompts beyond the widely used Cookie Theft (CT) picture description task, along with three diagnostic labels: dementia, MCI, and HC. 
The clinical relevance of \cognospeak{}, its data collection protocol, and our special feature-fusion and transfer learning architectures have been described in previous work \cite{pahar2025cognospeak, pahar2025mutlimodalfusion}. 
In this study, we extend these efforts by focusing specifically on the multilingual cohorts, who constitute the majority, within \cognospeak{} data to investigate whether standard AI architectures, such as ASR systems, large language models (LLMs), classifiers and regressors, exhibit systematic false positive biases favouring monolingual participants.
Furthermore, we employ qualitative analysis, such as \tfidf{}, to provide insights into the nature and sources of any identified biases.


\section{Data}
\label{sec:data}

\begin{table*}
    \centering
    \setlength{\leftmargini}{0.4cm}
    \caption{\textbf{Inclusion and exclusion criteria} of the multilingual participants at the community centres while they took part in \cognospeak{} assessments}
    \label{table:inclusion_exclusion}
    \begin{tabular}{| m{7cm} | m{7cm} |}
        \hline
        Inclusion & Exclusion \\
        \hline
        \begin{itemize} 
            \item Older than 50 years 
            \item Spoke more than one language
      \item Part of an ethnic minority community   
        \end{itemize}& 
        \begin{itemize} 
            \item Not able to complete \cognospeak{} assessments in English 
            \item Diagnosed with a neurodegenerative condition 
            \item Medical condition that would limit potential to take part in longer-term follow-up (including palliative diagnosis, like cancer, heart failure, COPD, etc.)
            \item Severe active psychiatric disease, including Schizophrenia and severe depression. (mild cases would be considered).

        \end{itemize}\\
        \hline
    \end{tabular}
\end{table*}

\begin{table*}[h]
    \centering
    \setlength{\tabcolsep}{2pt} 
    \renewcommand\arraystretch{1.3}
    \caption{\textbf{Dataset description.} The monolingual group consists of only English speakers, and the multilingual group contains speakers who speak English as an additional language to a primary language (mentioned in the `Language' column). Two healthy monolingual groups, one younger and the other older, are used for testing and comparison with the four multilingual groups to reduce age bias. Participants speaking Asian languages such as Urdu, Hindi, Punjabi, Mirpuri, Arabic, and English with West and South Yorkshire accents are noted as Asian (WY) and Asian (SY), respectively. Gender labels are \emph{F = female}, \emph{M = male}, along with 15 as undisclosed. Ages are listed along with the standard deviations. The audio column shows the total length of audio in hours. The `Whisper', `\WV{}' and `\Nemo{}' columns show the total number of words (in 1,000). The number of participants with available MMSE scores and manual transcripts is mentioned in the `MMSE' and `Trans' columns, respectively.  }
    \label{table:dataset}
    \begin{tabular}{c |c | c | c| c | c | c | c | c | c | c | c | c | c | c }
    
    \toprule
    \textbf{Evaluation} & \textbf{Language} & \textbf{Group} & \textbf{Diagnosis} & \textbf{\textit{N}} & \textbf{Age} & \textbf{M} & \textbf{F} & \textbf{Audio} & \textbf{SNR} & \textbf{Whisper} & \textbf{Wav2Vec2} & \textbf{\Nemo{}} & \textbf{MMSE} & \textbf{Trans}\\ 
    
    \midrule
    \multirow{3.5}{*}{Train} & \multirow{3.5}{*}{English (Train)} & \multirow{7}{*}{Monolingual} & Dementia  & 56 &  71.36$\pm$6.32  & 35  & 20 & 11.19 & -31.06 dB & 55.25 & 57.44 & 52.37 & 48 & 46 \\ 
    \cmidrule{4-15}
     & & & MCI & 150 & 70.17$\pm$6.04  & 78  & 70 & 32.93 & -30.91 dB & 197.19 & 205.67 & 187.2 & 119 & 113 \\
    \cmidrule{4-15}
     & & & HC & 768 & 66.53$\pm$6.27 & 273  & 488 & 143.68 & -32.50 dB & 947.24 & 993.42 & 928.53 & 7 & 56 \\

    \cmidrule{1-2}
    \cmidrule{4-15}
    \multirow{7}{*}{5-fold} & Younger English &  & HC & 163 &  48.44$\pm$4.17 & 41  & 119 & 28.71 & -31.60 dB & 190.49 & 198.90 & 189.33 & 5 & 11 \\ 
    \cmidrule{2-2}
    \cmidrule{4-15}
    \multirow{6}{*}{Train \&} & Older English & & HC & 56 & 82.68$\pm$2.98  & 32  & 23 & 12.52 & -32.04 dB & 75.39 & 78.27 & 70.92 & 44 & 22 \\
    
    \cmidrule{2-15}
    \multirow{5}{*}{Test} & Somali & \multirow{6}{*}{Multilingual} & HC & 54 & 41.04$\pm$13.51 & 26 & 28 & 9.35 & -30.09 dB & 51.68 & 53.65 & 49.01 & 45 & 50 \\
    \cmidrule{2-2}
    \cmidrule{4-15}
     & Chinese &  & HC & 47 & 55.98$\pm$10 & 17  & 30 & 7.92 & -31.44 dB & 37.3 & 40.33 & 37.53 & 44 & 37 \\
    \cmidrule{2-2}
    \cmidrule{4-15}
     & Asian (WY) &  & HC & 51 & 54.18$\pm$11.43 & 18  & 33 & 8.15 & -32.09 dB & 46.65 & 48.79 & 46.97 & 46 & 8\\
    \cmidrule{2-2}
    \cmidrule{4-15}
    & Asian (SY) &  & HC & 50 & 51.54$\pm$8.54 & 13  & 36 & 8.55 & -31.82 dB & 36.76 & 39.2 & 32.49 & 48 & 6 \\ 
    
    \midrule
    
    TOTAL & --- & --- & --- & 1395 & 63.32$\pm$11.43 & 533 & 847 & 263.05 & -32.08 dB & 1637.95 & 1715.67 & 1594.35 & 406 & 349 \\
    
    \bottomrule
\end{tabular}
\end{table*}

\subsection{Data collection}
\label{subsec:data_collect}

\cognospeak{} employs a virtual agent to administer 14 clinically validated, memory-probing questions, covering motivation ($\mathbf{Q_1}$), qualifying memory ($\mathbf{Q_2}$) \cite{weissberger2017diagnostic}, concern assessment ($\mathbf{Q_3}$) \cite{harrison2025may}, memory recall ($\mathbf{Q_4}$–$\mathbf{Q_8}$) \cite{ashford2008screening, lorentz2002brief}, cognitive functioning ($\mathbf{Q_9}$) \cite{cipriani2020daily}, fluency ($\mathbf{Q}_{10}$–$\mathbf{Q}_{11}$) \cite{vaughan2018semantic}, picture description ($\mathbf{Q}_{12}$–$\mathbf{Q}_{13}$) \cite{forbes2005detecting}, and reading tasks ($\mathbf{Q}_{14}$) \cite{noble2000oral} in a fixed order \cite{pahar2025mutlimodalfusion}.
Multilingual participants completed these assessments either at home or, with support from a research champion, in community centres, where South Asian populations represent 31\% of Bradford’s residents and 7\% in Sheffield \cite{cencus2021}. The inclusion and exclusion criteria are summarised in Table \ref{table:inclusion_exclusion}.

During data collection, several challenges were encountered. First, there was a scarcity of older ethnic minority participants with sufficient English proficiency to interact effectively with the \cognospeak{} agent. Second, technological barriers, such as unreliable internet connections at home or within community centres, limited some participants’ ability to complete the assessments. Finally, recording artefacts, including background noise or additional speech, were common because assessments conducted in communal spaces often required assistance from community or healthcare workers. 
Initially, recruitment focused on multilingual adults over the age of 50 years, to better align with the typical age range of patients referred to memory assessment services with functional cognitive disorder (FCD), MCI, or dementia \cite{ball2020dementia, vasileva2025diagnostic}. However, due to the relatively younger demographic profiles of some community centres, and limited English proficiency among older members, the inclusion criteria were later expanded to allow healthy adults of all ages.

Importantly, all multilingual participants with cognitive impairment lacked the English skills necessary to complete the \cognospeak{} assessments, resulting in the recruitment of only healthy individuals in these cohorts. In addition to \cognospeak{}, participants also completed pen-and-paper cognitive assessments, including the Montreal Cognitive Assessment (MoCA), which was subsequently converted to MMSE scores \cite{fasnacht2023conversion}.

\subsection{Data description}
\label{subsec:data_desc}

A total of 1,395 participants, representing both monolingual and multilingual groups in the UK, were included in this study, each completing 14 question prompts, as demonstrated in Table \ref{table:dataset}. Together, the dataset contains 263.05 hours of recorded speech, which was automatically transcribed using three ASR systems: Whisper-medium \cite{radford2022whisper}, \WV{} \cite{baevski2020wav2vec}, and NVIDIA \Nemo{} (NVIDIA Neural Modules 1.21.0) \cite{kuchaiev2019nemo}, yielding 1.64, 1.72, and 1.59 million words, respectively. To assess audio quality, we calculated the signal-to-noise ratio (SNR) between speech and pauses, with speech segments identified using Silero VAD, a lightweight and accurate voice activity detector \cite{Silero_VAD}. As shown in Table \ref{table:dataset}, the recordings achieved an average SNR of –32.08 dB, 
demonstrating the robustness of the \cognospeak{} recordings under real-world data collection conditions \cite{wu2018characteristics}.
Due to the high cost and time demands, manual transcripts were available for only 349 participants, which were then used to calculate WERs across language groups. Similarly, MoCA scores were available for a subset of 406 participants.

A two-tailed $t$-test at the 1\% level of significance ($p < 0.01$) confirmed that the length of audio and the number of words per assessment did not differ significantly across diagnostic groups. 
Initially, age was significantly different between cases (dementia and MCI) and controls (HC), particularly among multilingual participants, owing to the scarcity of older multilingual recruits (Section \ref{subsec:data_collect}). To mitigate this age bias in training, healthy monolingual participants were stratified into a younger group (40–55 years) and an older group (80+ years), together with four multilingual groups, which were designated as test cohorts, while the middle-aged group (55–80 years) was combined with dementia and MCI cohorts for training.
The final datasets are presented in Table \ref{table:dataset}, and no statistically significant differences were found between the younger English and the four multilingual test sets.


\section{Experimental setup}
\label{sec:ex_setup}

To investigate potential bias in AI-based speech screening systems, and in light of the absence of cognitively impaired participants in the multilingual group, we employed four complementary evaluation techniques across monolingual and multilingual speakers: (i) assessing ASR performance via word error rate (WER), (ii) classification-based analysis of FPR, (iii) regression of Mini-Mental State Examination (MMSE) scores, and (iv) qualitative linguistic analysis using \tfidf{}. 
These complementary analyses allow us to evaluate both predictive behaviour and disparities in model outputs across speaker groups.

The corresponding feature extraction, classifier training, and evaluation process are detailed below.   

\subsection{Feature Extraction and Classifier Training}
\label{subsec:feat_train}

Acoustic features, 
such as the extended Geneva Minimalistic Acoustic Parameter Set (\egemaps{} \cite{eyben2015geneva}) and The INTERSPEECH 2016 Computational Paralinguistics Challenge (\compare{} \cite{schuller2016interspeech, eyben2010opensmile}) features,
were extracted from the speech data, while linguistic features such as \tfidf{} and Bag-of-Words (\bow{}) were derived from the \Nemo{}-generated transcripts. These feature matrices from the audio and text domains were then fused to train a classifier (SVM) and a regressor (SVR), which had shown the best performance in our prior work \cite{pahar2025cognospeak, pahar2025mutlimodalfusion}. In addition, three LLMs (\bart{}, \roberta{}, and \distilbert{}) were fine-tuned and evaluated on the \Nemo{} transcripts; as their individual performances were comparable in \cite{pahar2025cognospeak}, we report the averaged results in Section \ref{subsec:classification}. 
We further note that our \cognospeak{} data is currently highly imbalanced (Table \ref{table:dataset}), and no data balancing technique apart from adjusting the classifier weights according to the class imbalance has been adopted.
These experiments allow us to examine predictive performance as well as potential disparities in model behaviour across monolingual and multilingual speaker groups.

\subsection{Evaluation}
\label{evaluation}

Table \ref{table:dataset} summarises that the training set comprises participants with 56 cases of dementia, 150 MCI and 768 as HC. 
The six test sets contain two monolingual English groups, one younger and the other older, and four multilingual speakers from four communities with two different accents, all healthy. 
This lack of positive cases (dementia or MCI) among multilingual participants posed a substantial challenge for standard classifier evaluation. To overcome this, the training dataset with three diagnostic labels was fixed, and a five-fold cross-validation scheme with strict no-participant overlap was applied to the remaining six datasets in Table \ref{table:dataset}. For example, in the Somali group, classifiers were trained on the fixed dataset plus four folds of Somali data, and tested on the remaining fold, with the process repeated across all five folds. 
As each of the six test sets contained only the HC group (a single class), standard multi-class metrics such as macro $F_1$-score were not applicable \cite{pahar2024nanocnn, christen2023review, paharliudatapaper}. Instead, we evaluated model behaviour using the FPR, defined as the proportion of healthy control participants incorrectly classified as cognitively impaired. Accuracy values are also reported for completeness. Final performances were averaged across folds, as described in Section \ref{sec:results}.
Given that all multilingual participants in the evaluation sets were healthy controls, the experimental design enables a direct analysis of false positive bias in speech-based AI screening systems.
To quantify disparities between speaker groups, we additionally report the relative risk ratio, defined as the ratio of the FPR for multilingual speakers to that of monolingual speakers.
For statistical analysis, a standard unpaired two-tailed $t$-test with a 5\% significance level was applied throughout Section \ref{sec:results} to compute $p$-values and assess the presence of bias across AI models. 


Importantly, multilingual data were incorporated into the training sets to improve model generalisation across speaker groups and to enable a more robust investigation of potential disparities between monolingual and multilingual participants.
This strategy allowed us to make the most effective use of limited multilingual data, while minimising the impact of the absence of cognitively impaired multilingual participants on our bias evaluation.


\section{Results}
\label{sec:results}

This section reports the results of the ASR evaluation, classification experiments, regression analysis, and bias analysis across monolingual and multilingual cohorts described in Table~\ref{table:dataset}.

\subsection{ASR Performance}
\label{subsec:wer}
\begin{figure*}[h]
  \centering 
  \includegraphics[width=0.7\linewidth]{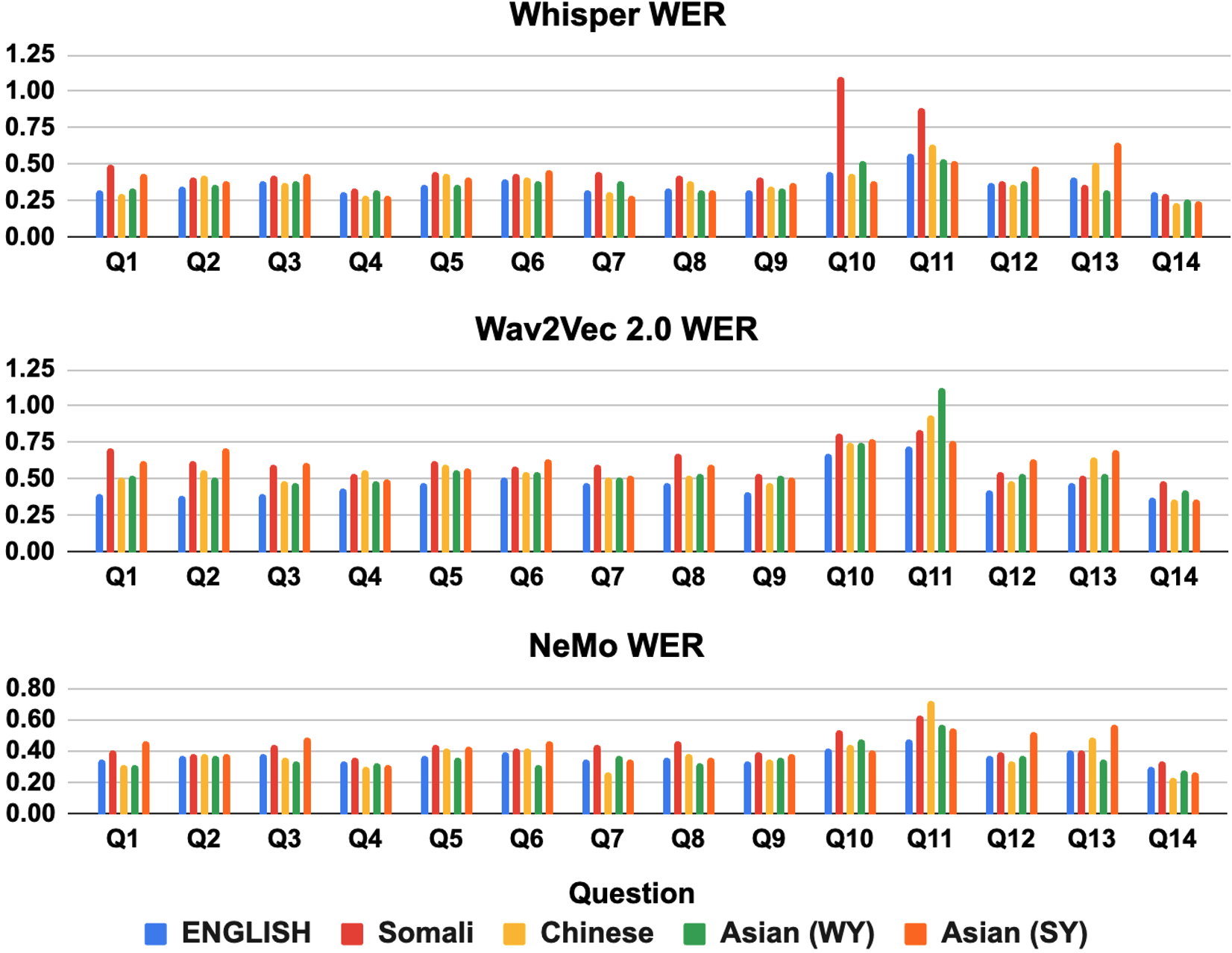} 
  \caption{ \textbf{Distribution of WERs across various question prompts for three ASRs.} The monolingual group is noted as `English'. It shows a trend of having a higher WER for the fluency tests ($\mathbf{Q}_{10}$, $\mathbf{Q}_{11}$). 
  Overall, Whisper and \WV{} have similar performance ($p$-value = 0.66), whereas \Nemo{}'s performance is significantly different ($p$-value =$< 0.0001$).
  }
  \label{fig:WER_ASR}
\end{figure*}

Automatic speech recognition (ASR) performance was evaluated using three off-the-shelf systems: Whisper, \WV{}, and \Nemo{}. 
The WERs were calculated between the manual transcripts (349 participants with manual transcriptions in Table \ref{table:dataset}) and ASR-generated transcripts, and are
shown in Figure \ref{fig:WER_ASR}, which 
shows a trend of lower performance of the ASRs for fluency tasks ($\mathbf{Q}_{10}$, $\mathbf{Q}_{11}$), and the corresponding $p$-values (all are $< 0.0001$) suggest that these performances are significantly lower, as a higher WER indicates a lower ASR performance.

For monolinguals, the $p$-values are 0.008, 0.896 and 0.005 for Whisper vs \WV{}, Whisper vs \Nemo{}, and \WV{} vs \Nemo{}, respectively; and they are much lower ($< 0.0001$, 0.69, $< 0.0001$) for multilingual speakers.
This suggests a larger performance gap between the three ASRs for multilingual speakers than for monolinguals.

The overall mean and standard deviation of WERs using all 14 questions for Whisper, \WV{} and \Nemo{} are 0.41$\pm$0.11, 0.57$\pm$0.15 and 0.40$\pm$0.11, respectively, as described in Table \ref{table:wer_summary}.
The $p$-values are $< 0.0001$, 0.67, and $< 0.0001$, for Whisper vs \WV{}, Whisper vs \Nemo{}, and \WV{} vs \Nemo{}, respectively; 
demonstrating the performances of Whisper and \Nemo{} are not significantly different, whereas the performance of \WV{} is significantly lower. 
It also indicates that overall \Nemo{} is the ASR of choice while transcribing speech to text automatically, due to its marginally better standard deviation over Whisper. 
Table \ref{table:wer_summary} also points out an overall worse performance of the Whisper and \WV{} for Somali multilingual speakers, whereas \Nemo{} has been consistent over all language groups. 


We have also experimentally discovered that although the WERs are marginally higher for the dementia group (monolingual) and the Somali group (multilingual), the WERs from all seven groups in Table \ref{table:wer_summary}
were not statistically significantly different from each other, as $p$-values were much higher, ranging between 0.2 and 0.7. 
The overall $p$-value of 0.1386 suggests no significant differences between the ASR performances of mono and multilingual English speakers across the UK.

\begin{table}[h]
    \centering
    \scriptsize
    \setlength{\tabcolsep}{2pt}
    \renewcommand\arraystretch{1}
    \caption{\textbf{Summary of WERs generated from three ASRs for mono and multilingual speakers}, showing \Nemo{} has achieved the lowest WER among three ASRs. An overall $p$-value of 0.1275 suggests the WERs are not significantly different between monolingual and multilingual speakers.}
    \label{table:wer_summary}
    \begin{tabular}{c | c  c  c  c  c  c  c  c }
    \toprule
    \multirow{2.5}{*}{\textbf{ASR}} & \multicolumn{3}{c|}{\textbf{Monolingual}} & \multicolumn{4}{c|}{\textbf{Multilingual}} & \multirow{2.5}{*}{\textbf{Avg}} \\
    \cmidrule{2-8}
     & \textbf{Dementia} & \textbf{MCI} & \multicolumn{1}{c|}{\textbf{HC}} & \textbf{Somali} & \textbf{Chinese} & \textbf{Asian(WY)} & \multicolumn{1}{c|}{\textbf{Asian(SY)}} & \\ 
    \midrule
    \textbf{Whisper} & \heatmapcell{0.41} & \heatmapcell{0.34} & \heatmapcell{0.39} & \heatmapcell{0.60} & \heatmapcell{0.41} & \heatmapcell{0.37} & \heatmapcell{0.41} & \heatmapcell{0.41} \\
    \textbf{W2V2} & \heatmapcell{0.51} & \heatmapcell{0.43} & \heatmapcell{0.49} & \heatmapcell{0.63} & \heatmapcell{0.57} & \heatmapcell{0.62} & \heatmapcell{0.61} & \heatmapcell{0.57} \\
    \textbf{\Nemo{}} & \heatmapcell{0.40} & \heatmapcell{0.34} & \heatmapcell{0.39} & \heatmapcell{0.43} & \heatmapcell{0.39} & \heatmapcell{0.37} & \heatmapcell{0.43} & \heatmapcell{0.40} \\
    \textbf{Avg} & \heatmapcell{0.44} & \heatmapcell{0.37} & \heatmapcell{0.42} & \heatmapcell{0.63} & \heatmapcell{0.46} & \heatmapcell{0.45} & \heatmapcell{0.48} & \heatmapcell{0.46} \\
    \bottomrule
    \end{tabular}
    \vspace{-5pt}
\end{table}




\subsection{Classification Performance}
\label{subsec:classification}




To explore better generalisation, we have used both \cognospeak{} and \dementiabank{} data \cite{becker1994natural} (one of the few publicly available dementia detection datasets) to train our classifiers on both cognitively impaired and healthy participants and then tested on the specially curated test set in Table \ref{table:dataset}. 

\begin{figure*}[h]
  \centering 
  \includegraphics[width=0.75\linewidth]{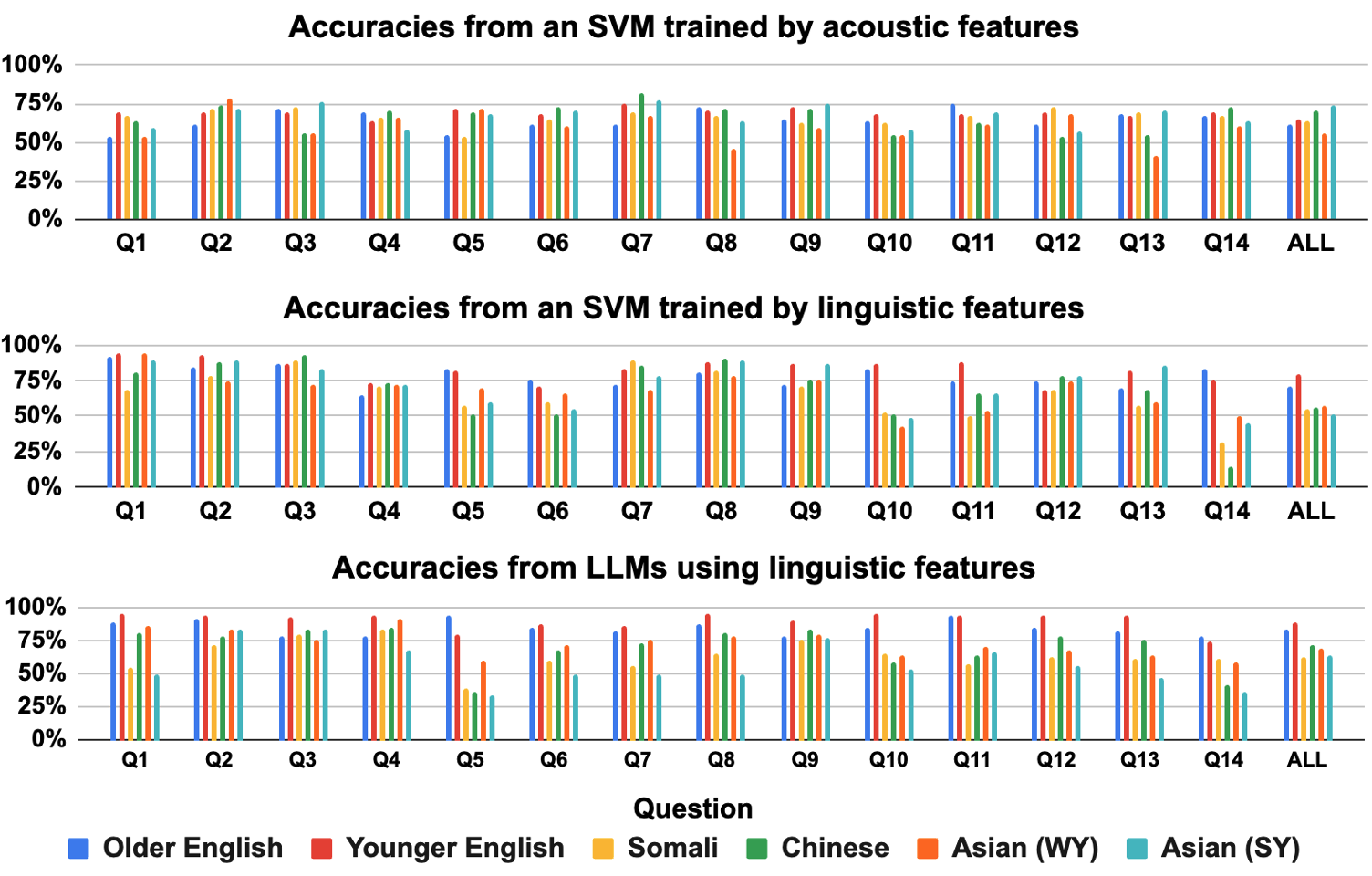} 
  \caption{ \textbf{Accuracies obtained from classifiers on predicting monolingual (Older and Younger English) and multilingual (Somali, Chinese and Asian with West Yorkshire and South Yorkshire accents) English speakers.} It shows that SVM trained by acoustic features have performed the worst, but is more consistent among mono and multilingual speakers. Whereas LLMs performed the best, but show a stronger bias, which is further confirmed by the $p$-values in Table \ref{table:accuracies_p_val}.}
  \label{fig:accuracies_ALL}
  \vspace{-5pt}
\end{figure*}

\begin{table}[h]
\centering
\setlength{\tabcolsep}{10pt} 
\renewcommand\arraystretch{1.2}
\caption{\textbf{Statistically significant $p$-values between monolingual and multilingual speakers.} Results indicate significant bias in memory ($Q_5, Q_6$), fluency ($Q_{10}, Q_{11}$), and reading tasks ($Q_{14}$) primarily when using linguistic features. Significance levels: $^{*}p<0.05$, $^{**}p<0.01$, $^{***}p<0.001$.}
\begin{tabular}{l | r r r}
    \toprule
    \textbf{Question} & \textbf{SVM+Audio} & \textbf{SVM+Text} & \textbf{LLM+Text} \\
    \midrule
    $Q_1$  & \sig{0.9428} & \sig{0.2975} & \sig{0.1836} \\
    $Q_2$  & \sig{0.0678} & \sig{0.3112} & \sig{0.0351} \\
    $Q_3$  & \sig{0.5558} & \sig{0.7788} & \sig{0.3516} \\
    $Q_4$  & \sig{0.6978} & \sig{0.3154} & \sig{0.7015} \\
    $Q_5$  & \sig{0.7488} & \sig{0.0114} & \sig{0.0111} \\
    $Q_6$  & \sig{0.6218} & \sig{0.0348} & \sig{0.0401} \\
    $Q_7$  & \sig{0.4781} & \sig{0.7157} & \sig{0.1144} \\
    $Q_8$  & \sig{0.3176} & \sig{0.8941} & \sig{0.0994} \\
    $Q_9$  & \sig{0.7749} & \sig{0.8177} & \sig{0.2941} \\
    $Q_{10}$ & \sig{0.0674} & \sig{0.0009} & \sig{0.0071} \\
    $Q_{11}$ & \sig{0.1339} & \sig{0.0410} & \sig{0.0011} \\
    $Q_{12}$ & \sig{0.7338} & \sig{0.3781} & \sig{0.0381} \\
    $Q_{13}$ & \sig{0.4366} & \sig{0.4539} & \sig{0.0616} \\
    $Q_{14}$ & \sig{0.6744} & \sig{0.0307} & \sig{0.0399} \\
    \midrule
    ALL & \sig{0.7214} & \sig{0.0088} & \sig{0.0019} \\
    \bottomrule
\end{tabular}
\label{table:accuracies_p_val}
\end{table}


\subsubsection{3-way classification using \cognospeak{} data}

The corresponding accuracies from the classifiers trained on the \cognospeak{} data are shown in Figure \ref{fig:accuracies_ALL}. 
The mean and standard deviation of the accuracies achieved from the SVM trained on the acoustic features are 0.67$\pm$0.09. 
This increases to 0.71$\pm$0.19, while using the linguistic features used to train the SVM, and a further 2\% increase (0.73$\pm$0.11) is observed while fine-tuning the LLMs with texts. 
This pattern is visually demonstrated in Figure \ref{fig:accuracies_ALL}. 

To effectively investigate the difference between the classification accuracies of the models (i.e., detecting bias) for monolingual and multilingual participants, 
the corresponding $p$-values are generated and outlined in Table \ref{table:accuracies_p_val}. 
While the model's classification accuracy did not significantly differ between language backgrounds when using only acoustic features, models using linguistic features generated significantly lower classification accuracies in the multilingual cohort, specifically in the memory, fluency, and reading tasks, consisting of ($\mathbf{Q}_5$, $\mathbf{Q}_6$), ($\mathbf{Q}_{10}$, $\mathbf{Q}_{11}$) and ($\mathbf{Q}_{14}$) respectively. 
This fact is further established while $p$-values such as 0.0002 (acoustic+SVM vs linguistics+SVM), $< 0.0001$ (acoustic+SVM vs linguistics+LLM) and 0.58 (linguistics+SVM vs linguistics+LLM) are achieved considering all questions and languages,
suggesting the presence of classifier bias towards monolinguals.

\begin{figure}[h]
  \centering 
  \includegraphics[width=0.65\linewidth]{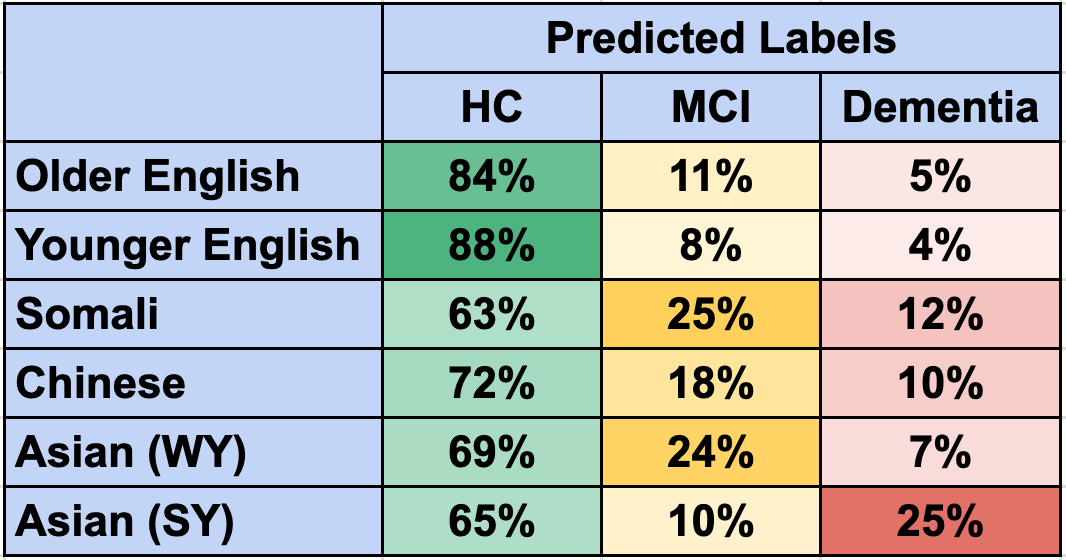}
  \caption{\textbf{Accuracies of the predicted labels from the 3-way classifiers} show 
    only 16\% (11\% MCI and 5\% dementia) have been misclassified exhibiting cognitive decline for the younger monolingual speakers, whereas those percentages rise to 35\% (10\% MCI and 25\% dementia) for Asian multilingual speakers with a South Yorkshire accent.}
  \label{fig:conf_mats}
  \vspace{-5pt}
\end{figure}

The prediction accuracies are also highlighted from the best-performing classifiers (LLMs) and shown in Figure \ref{fig:conf_mats}, which also sheds light on the severity of the misclassified diagnosis labels. 
The Figure \ref{fig:conf_mats} clearly shows that classifiers are more accurate in predicting healthy for monolingual speakers, and only a lesser percentage are classified as having dementia, which is a more severe form of cognitive decline. 
Whereas multilingual speakers, especially Asians with a South Yorkshire accent, are more likely to be misclassified as living with dementia than MCI.

\subsubsection{2-way classification using \dementiabank{} data}

\begin{figure*}[h]
  \centering 
  \includegraphics[width=0.7\linewidth]{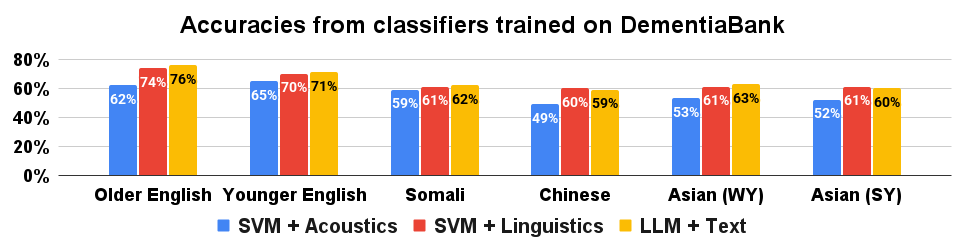} 
  \caption{ \textbf{Accuracies generated by the classifiers while trained on the \dementiabank{}} show a similar but stronger bias than the one found in Figure \ref{fig:accuracies_ALL}. Classifiers performed better for monolingual than multilingual speakers. }
  \label{fig:accuracies_DementiaBank}
  \vspace{-5pt}
\end{figure*}

Classifiers were also trained on the \dementiabank{} Pitt corpus, containing participants with only 2 diagnosis labels, and tested on the $\mathbf{Q}_{12}$ alone of \cognospeak{} data, as \dementiabank{} contained only the CT picture description task. 
The ethnicity of the participants in this dataset was unknown. 
The accuracies are shown in Figure \ref{fig:accuracies_DementiaBank}, showing a higher performance for the monolingual speakers (0.70$\pm$0.05) than multilingual speakers (0.58$\pm$0.04), constituting a \mbox{$p$-value} of $< 0.0001$, suggesting all classifiers using both acoustic and linguistic features are biased towards monolingual speakers in the UK. 

Unlike the pattern found in Table \ref{table:accuracies_p_val}, here, all three classifiers and feature combinations are significantly different at the 5\% level of significance, as the $p$-values of 0.04 (acoustics+SVM), 0.0009 (linguistics+SVM) and 0.004 (linguistics+LLM) are achieved between mono and multilingual speakers.
This indicates a strong bias towards monolingual speakers present in the \dementiabank{} dataset, whereas our \cognospeak{} data is comparatively slightly more generalisable to multilingual speakers in the UK.  

\subsubsection{FPR Analysis}

Since all test participants (Table \ref{table:dataset}) belong to the HC class, predictions of MCI or dementia represent false positives. The FPR, therefore, measures the proportion of healthy participants incorrectly classified as cognitively impaired by Equation \ref{eq:fpr}:

\begin{equation}
FPR = \frac{FP}{FP + TN}
\label{eq:fpr}
\end{equation}

Using the cohort sizes in Table~\ref{table:dataset} and the prediction proportions from Figure~\ref{fig:conf_mats}, the number of false positive predictions was estimated for each cohort in Table \ref{table:fpr}.

\begin{table*}[t]
\centering
\caption{False positive analysis across cohorts. The rightmost column reports the aggregated false positive rate (FPR) for each speaker group.}
\label{table:fpr}
\begin{tabular}{llccccc}
\toprule
Group & Cohort & N & FP & TN & FPR & Combined FPR \\
\midrule

\multirow{2}{*}{Monolingual}
& Older English   & 56  & 9  & 47  & 0.16 & \multirow{2}{*}{0.132} \\
& Younger English & 163 & 20 & 143 & 0.12 &  \\
\midrule

\multirow{4}{*}{Multilingual}
& Somali     & 54 & 20 & 34 & 0.37 & \multirow{4}{*}{0.332} \\
& Chinese    & 47 & 13 & 34 & 0.28 &  \\
& Asian (WY) & 51 & 16 & 35 & 0.31 &  \\
& Asian (SY) & 50 & 18 & 32 & 0.35 &  \\
\bottomrule

\end{tabular}
\label{table:fpr}
\end{table*}

The results reveal substantially higher FPR among multilingual speakers. Monolingual English cohorts show FPR between 12\% and 16\%, whereas multilingual groups exhibit rates ranging from 28\% to 37\%.
Aggregating across cohorts, monolingual speakers have an average FPR of approximately 13.2\%, compared with 33.2\% for multilingual speakers. This indicates that multilingual participants are more than twice as likely to be incorrectly classified as cognitively impaired.
Across the six cohorts, the false positive rates exhibited substantial variability (mean = 0.27, SD = 0.10), 
further highlighting disparities in model predictions across demographic groups.

\subsubsection{Odds Ratio and Risk Analysis}

To quantify the magnitude of the observed disparity, we compute the odds ratio (OR) comparing false positive outcomes between multilingual and monolingual speakers using Equaltion \ref{eq:oddsRatio}. 

\begin{equation}
OR =
\frac{FP_{multi}/TN_{multi}}{FP_{mono}/TN_{mono}}
\label{eq:oddsRatio}
\end{equation}

where $FP_{mono}=29$, $FP_{multi}=67$, $TN_{multi}=135$, and $TN_{mono}=190$, 
using the aggregated counts derived from Table~\ref{table:fpr}. 

This yields an odds ratio of $OR = 3.25$,
This yields an odds ratio of $OR = 3.25$, indicating that the odds of receiving a false positive classification are more than three times higher for multilingual speakers than for monolingual speakers.

To quantify the uncertainty of this estimate, we computed the 95\% confidence interval using the log odds ratio method:

\[
CI_{95\%} = [2.01, 5.26]
\]

Because the confidence interval does not include 1.0, the effect is statistically significant and indicates a substantial disparity between groups \cite{bland2000odds, altman1990practical}.

While the odds ratio quantifies the strength of association between speaker group and misclassification, the relative risk provides a more intuitive interpretation of the increase in false positive probability.
Thus, in addition to the odds ratio, we also compute the relative risk (risk ratio) using Equation \ref{eq:Riskratio}. 

\begin{equation}
RR =
\frac{P(FP | multilingual)}{P(FP | monolingual)}
\label{eq:Riskratio}
\end{equation}

Using the observed FPR, $RR = \frac{0.332}{0.132} \approx 2.52$. 
This indicates that multilingual participants are approximately 2.5 times more likely to receive a false positive prediction than monolingual participants.

\subsubsection{Chi-Square Significance Testing}

To evaluate whether the observed differences in prediction outcomes are statistically significant, we performed a chi-square test of independence comparing prediction outcomes (correct HC vs.\ false positive) between monolingual and multilingual speakers.
The contingency table is defined in Table \ref{table:contingency}.

\begin{table}[t]
\centering
\caption{Contingency table showing correct classifications and false positive predictions for monolingual and multilingual healthy control (HC) participants.}
\label{table:contingency}
\begin{tabular}{lcc}
\toprule
Group & Correct HC & False Positive \\
\midrule
Monolingual   & 190 & 29 \\
Multilingual  & 135 & 67 \\
\bottomrule
\end{tabular}
\label{table:contingency}
\end{table}

The chi-square statistic was calculated as $\chi^2 = 22.31$, with one degree of freedom.
The resulting p-value is $p < 0.0001$, 
indicating a highly significant association between speaker group and classification outcome. These results demonstrate that multilingual speakers are significantly more likely to be misclassified as cognitively impaired by the model.

\subsubsection{Fairness Metric Analysis}
To further quantify disparities in model behaviour across demographic groups, we evaluate fairness using standard classification fairness metrics commonly used in machine learning bias analysis \cite{barocas2023fairness}.

\textbf{FPR Difference: }
The FPR difference measures the disparity in misclassification rates between demographic groups using Equation \ref{eq:fprDiff}. 

\begin{equation}
FPR_{diff} = FPR_{multi} - FPR_{mono}
\label{eq:fprDiff}
\end{equation}

Thus, the observed FPR ($FPR_{mono} = 0.132, FPR_{multi} = 0.332$), yields $FPR_{diff} = 0.20$, 
indicating that multilingual speakers experience a 20 percentage point higher FPR than monolingual speakers. Such a disparity is substantial in clinical screening contexts, where false positives may trigger unnecessary diagnostic procedures or cause undue anxiety for patients.







\textbf{Statistical Effect Size: }
To quantify the magnitude of the association between speaker group and classification outcome, we compute the phi coefficient derived from the chi-square statistic:

\[
\phi = \sqrt{\frac{\chi^2}{N}}
\]

Using the previously computed $\chi^2 = 22.31$ and total evaluation sample size $N = 421$, we obtain $\phi = 0.23$.

According to conventional interpretations, this represents a moderate effect size, indicating a meaningful association between multilingual status and classification outcomes.

Together with the odds ratio (3.25) and statistically significant chi-square result ($p < 0.0001$), these fairness metrics provide strong evidence that the classification system exhibits systematic disparities across speaker groups.







\subsection{Regression Analysis}
\label{subsec:regress}

\begin{figure*}[t]
  \centering 
  \includegraphics[width=0.8\linewidth]{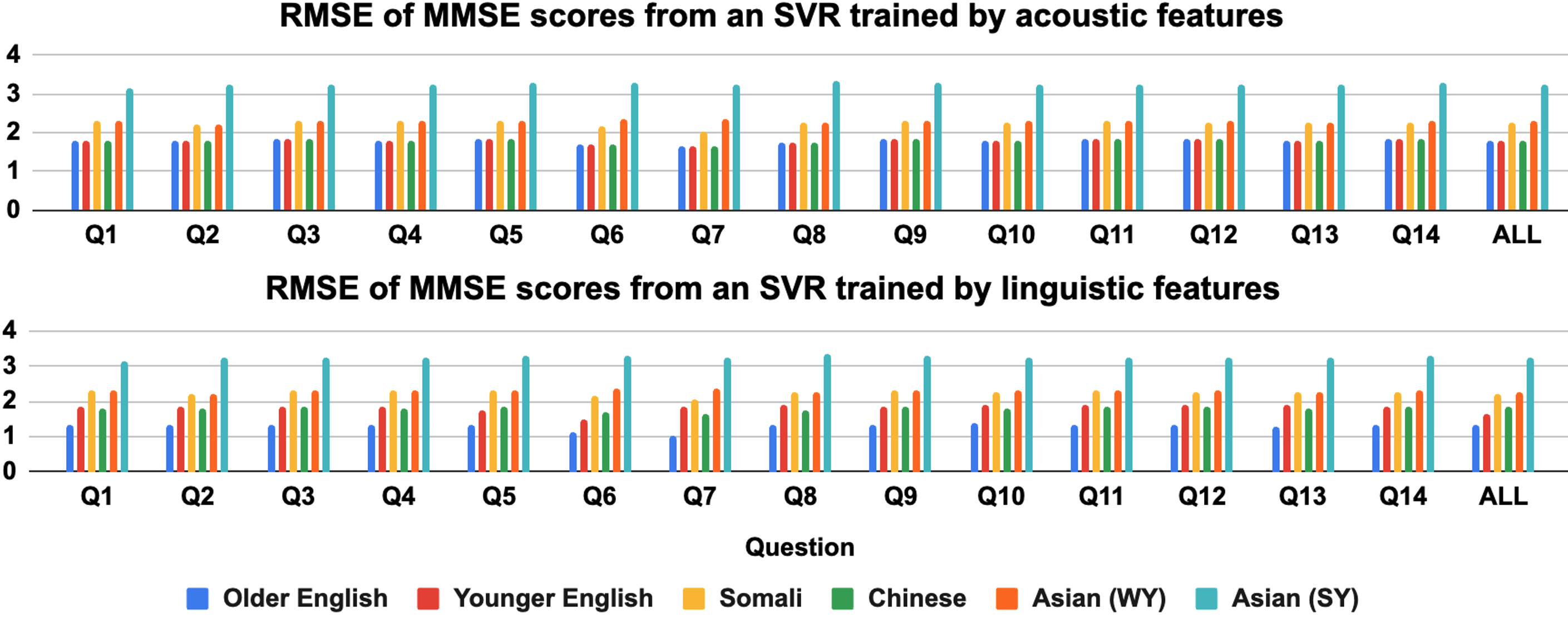} 
  \caption{ \textbf{RMSE-MMSE scores from an SVR trained on the linguistic features} showing lower RMSE (variations) among MMSE scores for monolingual than multilingual speakers. }
  \label{fig:RMSE_vals}
  \vspace{-10pt}
\end{figure*}

The MMSE scores were available for 406 participants in Table \ref{table:dataset}, and this screening test score has been used to calculate the Root Mean Squared Error (RMSE) values between the actual and predicted scores, after the SVR regressor was trained on both acoustic and linguistic features. 
The RMSE-MMSE scores for each question prompt are presented in Figure \ref{fig:RMSE_vals}, which shows lower MMSE scores in general for monolingual speakers for both feature sets. 
To evaluate whether the observed disparity in prediction error is statistically significant, we performed statistical testing across the RMSE-MMSE scores obtained for the 14 conversational prompts. The resulting test confirms a highly significant difference between monolingual and multilingual prediction errors
($p < 0.001$)
indicating that the higher prediction errors observed for multilingual speakers are unlikely to arise from random variation alone.

\subsubsection{Group-Level RMSE Disparity}
\label{subsec:rmse_group}

To quantify prediction disparities between demographic groups, we compare the average RMSE of MMSE predictions for monolingual and multilingual speakers using the aggregated results across the 14 conversational prompts.

The monolingual cohorts (Older English and Younger English) exhibit RMSE values of 1.34 and 1.65 respectively, yielding an average prediction error of

\[
RMSE_{mono} = \frac{1.34 + 1.65}{2} = 1.50
\]

In contrast, the multilingual cohorts (Somali, Chinese, Asian West Yorkshire, and Asian South Yorkshire) show RMSE values of 2.23, 1.82, 2.28, and 3.24, producing an average error of

\[
RMSE_{multi} =
\frac{2.23 + 1.82 + 2.28 + 3.24}{4} = 2.39
\]

This corresponds to an absolute error difference of

\[
\Delta RMSE = RMSE_{multi} - RMSE_{mono} = 0.89
\]

indicating that MMSE predictions for multilingual speakers exhibit nearly one additional point of prediction error compared with monolingual speakers.

\subsubsection{RMSE Ratio}
\label{subsec:rmse_ratio}

To further quantify the magnitude of this disparity, we compute the RMSE ratio between multilingual and monolingual speakers in Equaltion \ref{eq:RMSEratio}. 

\begin{equation}
RMSE_{ratio} =
\frac{RMSE_{multi}}{RMSE_{mono}}
\label{eq:RMSEratio}
\end{equation}

Substituting the observed values, we get $RMSE_{ratio} = \frac{2.39}{1.50} \approx 1.59$. 
This result indicates that prediction errors for multilingual speakers are approximately 59\% larger than those observed for monolingual speakers. Such a difference suggests that the regression model exhibits substantially lower predictive accuracy when applied to multilingual speech data.

\subsubsection{Performance Variability Across Cohorts}
\label{subsec:rmse_variability}

In addition to group-level disparities, we analyse the variability of prediction performance across individual cohorts. Considering the RMSE values across all six cohorts ($1.34,\ 1.65,\ 2.23,\ 1.82,\ 2.28,\ 3.24$), 
the mean RMSE is approximately $2.09$, with a standard deviation of $ 0.66$. 




This relatively large standard deviation indicates substantial variability in regression performance across demographic groups. The largest performance gap occurs between the Older English cohort ($RMSE=1.34$) and the Asian South Yorkshire cohort ($RMSE=3.24$), corresponding to a difference of $1.90$ RMSE-MMSE points. This gap highlights a pronounced disparity in prediction accuracy across language groups.

\subsubsection{Feature-Type Disparity Analysis}
\label{subsec:feature_disparity}

We further compare regression performance obtained using linguistic and acoustic features. For monolingual speakers, linguistic features yield an average RMSE of 1.50, compared with 1.78 when using acoustic features alone. This indicates that linguistic representations substantially improve prediction accuracy for monolingual speech.

However, for multilingual speakers the corresponding RMSE values are 2.39 (linguistic) and 2.40 (acoustic), indicating that linguistic features provide little improvement for multilingual data.

This asymmetric benefit suggests that language-dependent representations may capture patterns that are more consistent in monolingual speech, while failing to generalise effectively to multilingual speakers whose linguistic structures and vocabulary usage may differ substantially from those observed in the training data.

\subsubsection{Error Disparity Index}

Finally, we compute an error disparity index defined as the relative difference between multilingual and monolingual prediction errors in Equaliton \ref{eq:EDI}. 

\begin{equation}
EDI =
\frac{RMSE_{multi} - RMSE_{mono}}{RMSE_{mono}}
\label{eq:EDI}
\end{equation}

Substituting the observed values, $EDI = \frac{2.39 - 1.50}{1.50} \approx 0.59$, indicating a 59\% increase in prediction error for multilingual speakers relative to monolingual speakers. Such a disparity highlights the need for improved modelling approaches that are robust to linguistic diversity.

\subsection{Unified Fairness Analysis Across Classification and Regression}

\begin{figure}[t]
\centering
\includegraphics[width=0.99\linewidth]{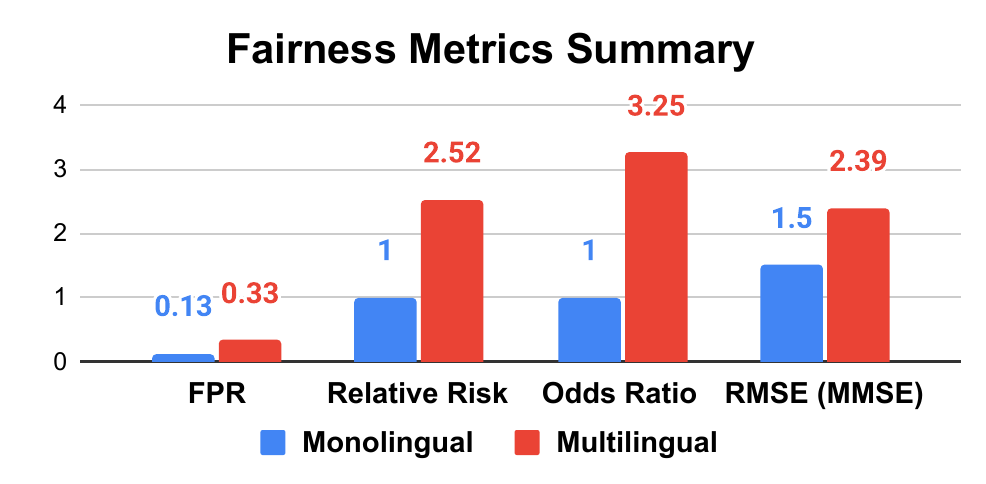}
\caption{\textbf{Unified fairness analysis across classification and regression tasks.}
Comparison of key performance metrics for monolingual and multilingual speakers. Multilingual participants exhibit higher false positive rates (FPR) in classification as well as larger prediction errors in MMSE regression. The consistency of these disparities across different evaluation metrics highlights systematic performance differences between demographic groups.}
\label{fig:fairness_summary}
\end{figure}

To provide a unified view of model behaviour across demographic groups, we summarise the principal fairness metrics derived from both classification and regression tasks.
Figure \ref{fig:fairness_summary} compares monolingual and multilingual participants across four key measures: FPR, odds ratio, relative risk, and RMSE of MMSE prediction. These metrics capture different aspects of model behaviour, including classification bias and regression accuracy disparities.

For the classification task, multilingual speakers exhibit a substantially higher false positive rate ($FPR=0.332$) compared with monolingual speakers ($FPR=0.132$). This disparity corresponds to an odds ratio of $OR=3.25$ and a relative risk of $RR=2.52$, indicating that multilingual participants are significantly more likely to be incorrectly classified as cognitively impaired.

For the regression task, multilingual speakers also experience larger prediction errors in MMSE estimation. The average RMSE for multilingual speakers is $2.39$, compared with $1.50$ for monolingual speakers. This represents a relative increase of approximately $59\%$ in prediction error.

Taken together, these results demonstrate a consistent pattern across both classification and regression analyses, suggesting that speech-based AI models trained primarily on monolingual data may exhibit systematic performance disparities when applied to multilingual populations.

\subsection{Explainable feature analysis: \tfidf{} }
\label{subsec:qualanalysis}

\begin{figure*}[t]
\centering     
\subfigure[\tfidf{} metric for \textbf{dementia}]{\label{fig:TDIDF_dementia}\includegraphics[width=0.32\linewidth]{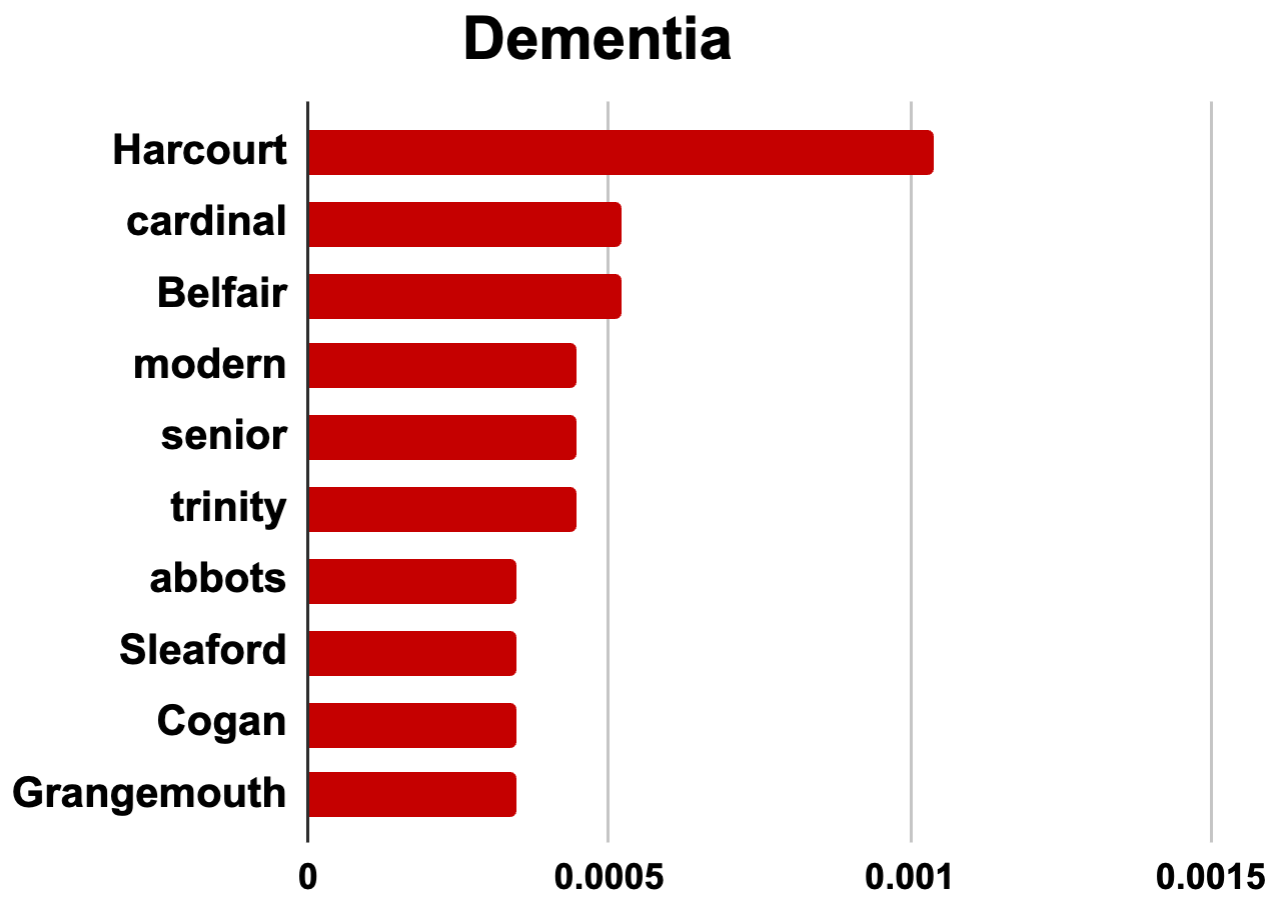}}
\subfigure[\tfidf{} metric for \textbf{HC}]{\label{fig:TFIDF_HC}\includegraphics[width=0.32\linewidth]{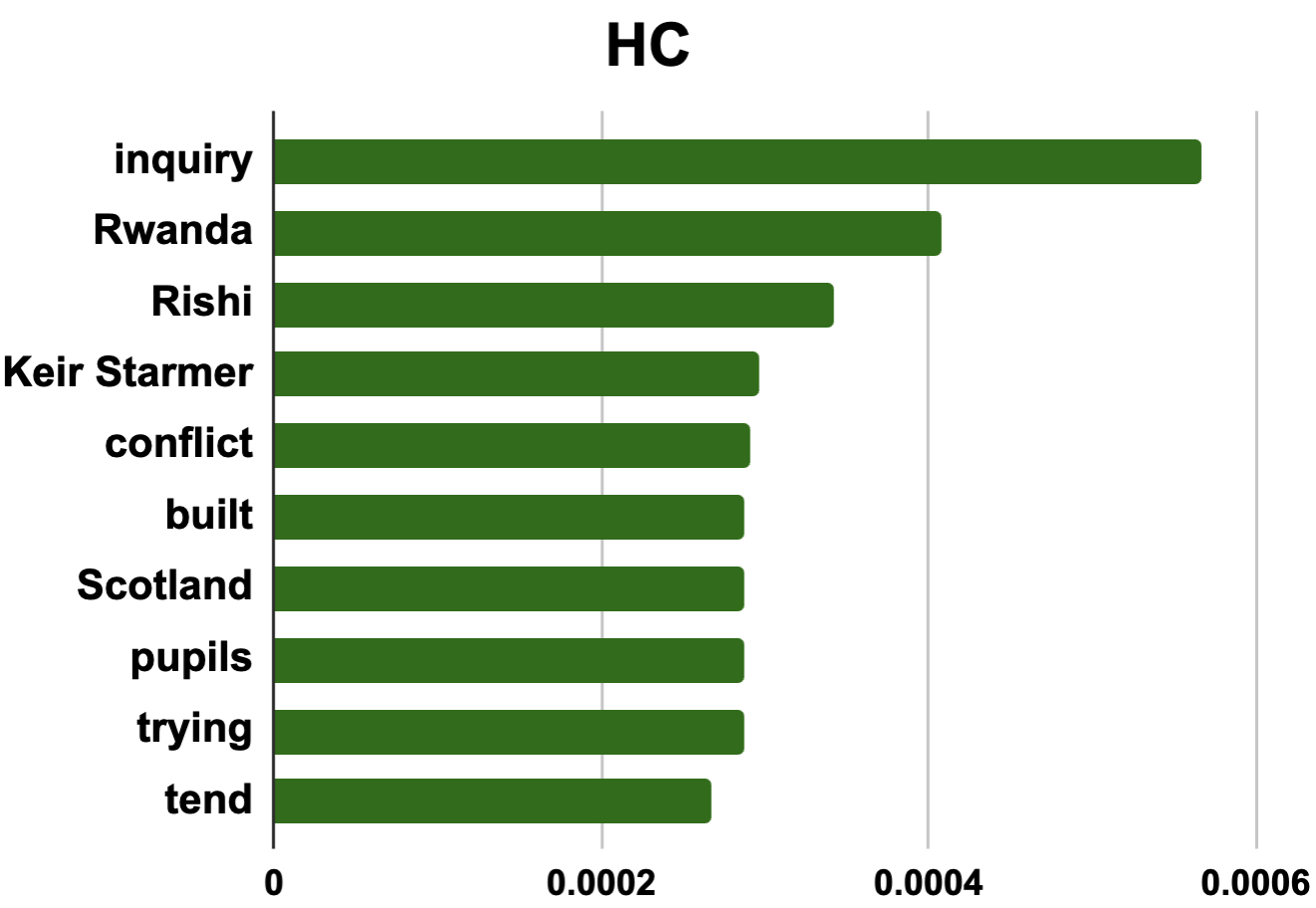}}
\subfigure[\tfidf{} metric for \textbf{multilingual South Asians}]{\label{fig:TFIDF_mult}\includegraphics[width=0.32\linewidth]{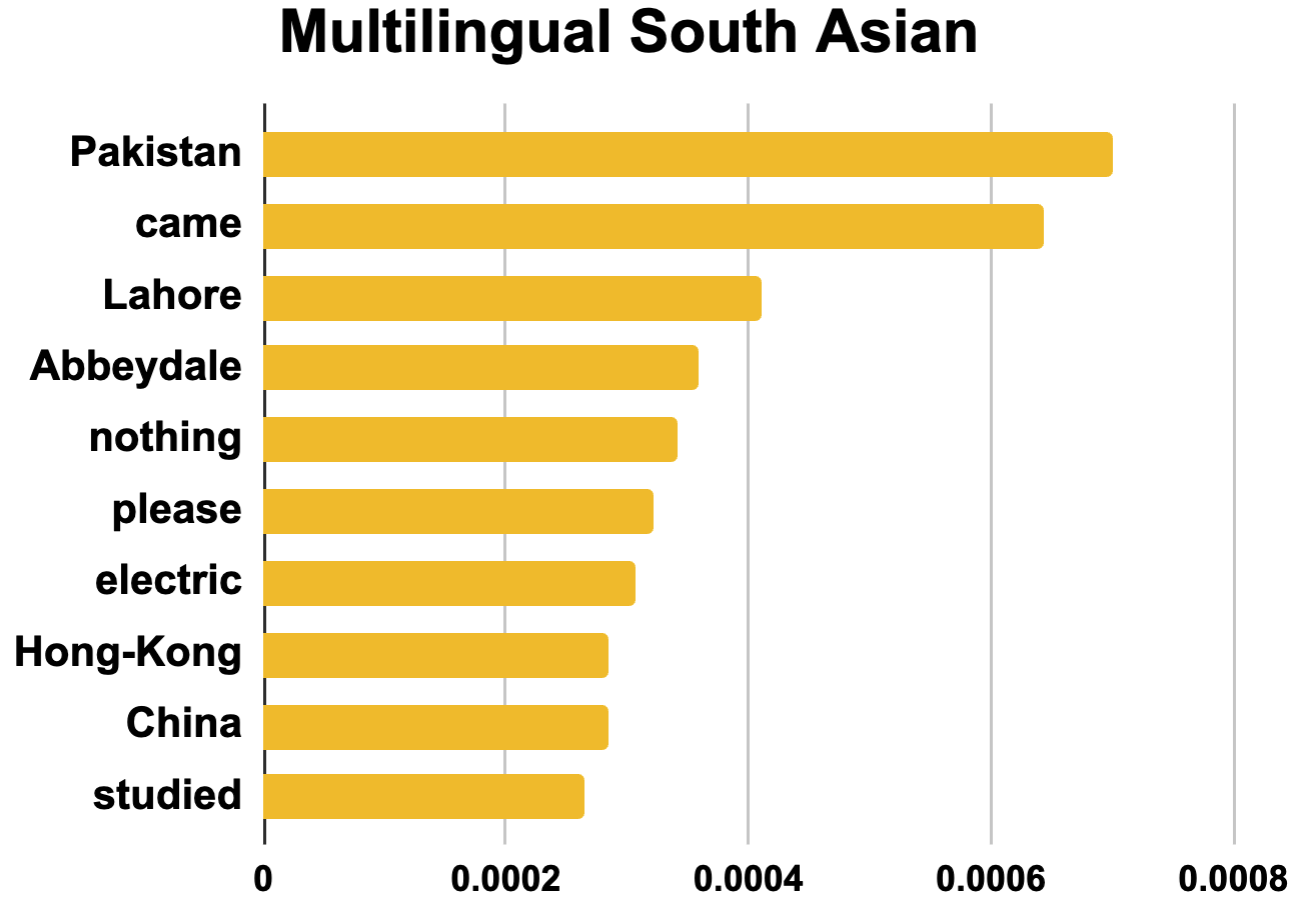}}
\caption{\textbf{Top-ten \tfidf{} metrics for the dementia, HC and multilingual groups while commenting on the memory tasks (`recent news' and `school-life')} shows that places like countries and towns are the most important words among these groups. Healthy participants also mention famous public figures like British prime ministers, and multilinguals are more likely to mention their place or country of origin.}
\label{fig:word_clouds__TFIDF}
\end{figure*}

To further investigate the data and classifier performance, the \tfidf{} metrics, used as the linguistics features, for the memory tasks ($\mathbf{Q}_{5}$, asking `what has been in the news recently' and $\mathbf{Q}_{6}$, asking `which school did you go to and when did you leave') between dementia, HC and South Asian multilingual groups (as this group has the highest percentage of participants predicted as living with dementia, the more severe type of cognitive decline in Figure \ref{fig:conf_mats}), and presented in Figure \ref{fig:word_clouds__TFIDF}. 
The top-10 \tfidf{} features indicate that words such as places are the most important features among these three groups. 
Participants living with dementia mention names of places (Figure \ref{fig:TDIDF_dementia}), whereas healthy participants also mention names of famous public figures such as British prime ministers (Figure \ref{fig:TFIDF_HC}). 
Moreover, multilingual South Asian speakers are likely to mention their countries and places of origin, such as \textit{Pakistan} and \textit{Lahore}, in Figure \ref{fig:TFIDF_mult}.


The x-axis of Figure \ref{fig:word_clouds__TFIDF} highlights subtle yet noteworthy differences across groups. 
For instance, \textit{Harcourt}, a UK locality, achieves the highest \tfidf{} score of 0.001 within the dementia corpus. 
By contrast, \textit{inquiry} and \textit{Pakistan} record the highest \tfidf{} values among the HC (0.00058) and South Asian multilingual (0.00065) corpora, respectively, both lower than the scores in the dementia group. 
This suggests that the dementia corpus contains more distinctive and comparatively more influential words within its top-10 \tfidf{} terms than the other two corpora.

\section{Discussion}
\label{sec:discussion}

We implemented four complementary approaches to investigate potential bias in AI models, with results summarised in Section \ref{sec:results}. When evaluating ASR performance using WER, we found that \Nemo{} performed marginally better than Whisper, while both substantially outperformed \WV{}. Unpaired two-tailed $t$-tests confirmed that WERs did not differ significantly between monolingual and multilingual participants, suggesting that ASRs exhibit limited bias and perform comparably well across groups. However, ASR performance declined considerably on fluency tasks, where speech was less spontaneous, a finding consistent with prior studies \cite{jain2020contextual, dingliwal2023personalization} and one that warrants further investigation. While fine-tuning could reduce WERs, we intentionally refrained in this pilot study to first assess the inherent performance and bias of off-the-shelf ASR models.

Classifier analysis revealed that models leveraging linguistic features achieved higher accuracy but displayed greater bias towards monolingual speakers, particularly in memory, fluency, and reading tasks. FPR analysis demonstrated substantial disparities across cohorts: monolingual English speakers exhibited rates between 12\% and 16\%, whereas multilingual speakers showed markedly higher rates ranging from 28\% to 37\%. Aggregated analysis indicated that multilingual participants were approximately 2.5 times more likely to receive a false positive classification than monolingual speakers (relative risk = 2.52), with an odds ratio of 3.25 (95\% CI [2.01, 5.26]). A chi-square test confirmed that this difference was statistically significant ($\chi^2 = 22.31$, $p < 0.0001$), with a moderate effect size ($\phi = 0.23$). Accent-related variation also contributed to classification disparities, with speakers exhibiting a South Yorkshire accent more frequently misclassified as experiencing more severe cognitive decline. Cross-dataset experiments further revealed that bias was present in both \cognospeak{} and \dementiabank{} datasets, although the magnitude differed: three-class classifiers trained on acoustic features from \cognospeak{} showed no detectable bias, whereas two-class classifiers trained on \dementiabank{} data exhibited clear disparities. These findings suggest that multilingual speakers face systematically higher FPR in current speech-based cognitive screening models.


Building on the analyses of ASR performance, classifier bias, and regression errors, a comprehensive fairness evaluation highlights systematic disparities affecting multilingual speakers. Multilingual participants exhibit substantially higher false positive rates (FPR = 0.332 vs. 0.132 for monolinguals), corresponding to a relative risk of 2.52 and an odds ratio of 3.25 (95\% CI [2.01, 5.26]). Regression of MMSE scores further demonstrates increased prediction errors for multilingual participants (average RMSE = 2.39 vs. 1.50 for monolinguals), representing a relative increase of approximately 59\%. 
Complementary analysis using \tfidf{} features confirms that cultural and linguistic differences, such as references to geographic locations or public figures, such as British prime ministers, may contribute to these disparities. Together, these results indicate that current speech-based AI models systematically misclassify multilingual healthy participants and produce less accurate cognitive predictions, underscoring the need for bias mitigation strategies and more inclusive model development to ensure equitable performance across diverse linguistic and ethnic populations in the UK.


\section{Limitations}
\label{sec:limitations}

This study has several limitations that should be considered when interpreting the results.

First, due to the absence of patients across the community centres involved, we were unable to recruit cognitively impaired multilingual participants. As a result, the multilingual cohorts consisted only of healthy control participants. Although our evaluation strategy was designed to mitigate this constraint through cross-validation and bias analysis, the absence of impaired multilingual speakers limits the ability to directly measure diagnostic sensitivity within these populations.

Second, the multilingual dataset represents a limited set of linguistic and cultural groups, primarily Somali, Chinese, and South Asian communities from Sheffield and Bradford. While these populations reflect important ethnic minority groups in the UK, they do not capture the full linguistic diversity of multilingual speakers across the country.

Third, manual transcripts and cognitive assessment scores were available only for a subset of participants due to the time and cost associated with annotation and clinical assessment. Although this subset was sufficient for statistical analysis, larger annotated datasets would allow more comprehensive investigation of linguistic bias and model behaviour.

Fourth, the models evaluated in this study were trained primarily on English speech data and publicly available datasets such as DementiaBank, which may themselves contain demographic and linguistic biases. Consequently, the observed disparities may partly reflect biases inherent in existing training resources rather than solely the modelling approaches.

Finally, this work represents a pilot investigation into fairness in speech-based cognitive screening systems. Future research should incorporate larger and more diverse multilingual cohorts, include cognitively impaired multilingual participants, and explore bias mitigation strategies such as domain adaptation, accent-aware modelling, and fairness-aware training methods.

\section{Ethics and Responsible AI Considerations}
\label{sec:ethics}

This study was conducted in accordance with ethical guidelines for research involving human participants. 
Ethical approval for the collection of the data analysed in this study was granted by the NRES Committee South West-Central Bristol (REC number 16/LO/0737).

All participants provided informed consent prior to taking part in the \cognospeak{} assessments, and data collection procedures were approved by the relevant institutional ethics committee. Particular attention was given to recruiting participants from underrepresented ethnic minority communities while ensuring voluntary participation and confidentiality.

Given the growing use of artificial intelligence in healthcare decision support systems, it is essential to evaluate potential biases that may arise when models are deployed across diverse populations. In this study, we explicitly examined disparities in model performance between monolingual and multilingual speakers to assess the trustworthiness of speech-based cognitive screening systems. Our results highlight the importance of developing AI models trained on representative datasets that reflect linguistic, cultural, and demographic diversity.

We emphasise that the AI models evaluated in this work are intended to support, rather than replace, clinical judgement. Automated predictions should therefore be interpreted with caution and used as complementary tools within broader clinical assessment frameworks. Future work will focus on improving the fairness, transparency, and robustness of these systems to ensure responsible deployment in real-world healthcare settings.


\section{Conclusion}
\label{sec:conclusion}

This study investigated false positive bias in speech-based artificial intelligence systems used for cognitive screening among multilingual English speakers in the United Kingdom (UK). Using acoustic and linguistic speech features extracted from over 263 hours of conversational assessments collected from 1,395 participants, we evaluated whether current AI models perform equitably across monolingual and multilingual populations.

Analysis of three off-the-shelf automatic speech recognition (ASR) systems (Whisper, \WV{}, and \Nemo{}) showed no statistically significant differences in word error rate between monolingual and multilingual speakers, indicating that transcription quality alone does not explain downstream disparities. However, classification models trained on linguistic features exhibited clear differences across language groups. Multilingual participants were consistently more likely to be misclassified as cognitively impaired, particularly in memory, fluency, and reading tasks.


FPR and regression analyses revealed substantial disparities across cohorts. Monolingual English speakers exhibited FPRs between 12\% and 16\%, whereas multilingual participants showed markedly higher rates ranging from 28\% to 37\%, corresponding to a false positive rate ratio of 2.52, an odds ratio of 3.25 (95\% CI [2.01, 5.26]), and a moderate effect size ($\phi = 0.23$). Regression of MMSE scores further indicated that prediction errors were approximately 59\% higher for multilingual participants compared with monolinguals, reinforcing the presence of systematic bias. Accent-related variation also contributed to disparities, with South Yorkshire speakers more frequently misclassified as exhibiting more severe cognitive decline. Collectively, these findings demonstrate that current speech-based AI models systematically misclassify multilingual healthy participants and produce less accurate cognitive predictions, highlighting significant fairness concerns for clinical deployment in diverse UK populations.


Future work will focus on improving the fairness and generalisability of speech-based cognitive screening systems by recruiting cognitively impaired multilingual participants, developing bias-mitigation strategies, and expanding explainable linguistic analyses. Incorporating culturally appropriate assessment tools, such as the Rowland Universal Dementia Assessment Scale (RUDAS), the Multicultural Cognitive Examination (MCE), and the Addenbrooke’s Cognitive Examination (ACE), may further improve equitable cognitive screening across diverse populations.



\section{Acknowledgement}
\label{sec:acknldge}

We acknowledge the support of NHS clinicians, who recruited the participants, and TherapyBox, which provided the data collection front-end app for CognoMemory, formerly CognoSpeak assessments. This research was partly funded by the NIHR Sheffield Biomedical Research Centre (BRC), and the NIHR202911 award under the NIHR i4i programme. The views expressed are those of the authors and not necessarily those of the NHS, the NIHR or the Department of Health and Social Care (DHSC). 
Ethical approval for the collection of the data analysed in this study was granted by the NRES Committee South West-Central Bristol (REC number 16/LO/0737). 
For the purpose of open access, the author has applied a Creative Commons Attribution (CC BY) licence to any Author Accepted Manuscript version arising. 
All participants took part in this study voluntarily and provided their consent to use their data for research purpose.
The data are not publicly available due to privacy and ethical restrictions; however, a subset of data with appropriate consents has been shared with the research community \cite{tao2025PROCESS, pahar2026PROCESS2}.


\vspace{20pt}


\bibliographystyle{IEEEtran}
\bibliography{mybib}

\end{document}